\documentclass[sigconf,nonacm]{acmart}

\renewcommand\footnotetextcopyrightpermission[1]{}
\AtBeginDocument{%
  }

\usepackage{booktabs,multirow,makecell}
\usepackage[table]{xcolor}
\usepackage{graphicx}
\usepackage{subcaption}   
\usepackage{enumitem}
\usepackage[most]{tcolorbox}    
\usepackage{xcolor}

\definecolor{wine}{RGB}{180,20,80}

\usepackage{algpseudocode}
\usepackage[ruled,vlined,linesnumbered]{algorithm2e} 
\usepackage{caption} 
\usepackage{wrapfig}

\usepackage{url}
\usepackage{booktabs}
\usepackage{multirow}
\usepackage{xcolor}

\usepackage{hyperref}     

\begin{document}
\pagestyle{plain}
\title{A Bridge from Audio to Video: Phoneme-Viseme Alignment Allows Every Face to Speak Multiple Languages}

\author{
Zibo Su,
Kun Wei,
Jiahua Li,
Jing Kong,
Xu Yang,
Cheng Deng\\
Xidian University\\
\textcolor{wine}{\url{https://tking-su.github.io/MuEx/}}
}

\renewcommand{\shortauthors}{}
\begin{abstract}
    Speech-driven talking face synthesis (TFS) focuses on generating lifelike facial animations from speech input. 
    Current TFS models perform well in English but unsatisfactorily in non-English languages, producing wrong mouth shapes and rigid facial expressions. 
    The terrible performance is  mainly caused by the English-dominated training datasets and the lack of cross-language generalization abilities.
    Thus, we propose Multilingual Experts \textbf{(MuEx)}, a novel framework featuring a Phoneme-Guided Mixture-of-Experts \textbf{(PG-MoE)} architecture that employs phonemes and visemes as universal intermediaries to bridge the gap between audio and video modalities, achieving lifelike multilingual TFS.
    To alleviate the influence of linguistic differences and dataset bias, we extract speech and visual features as phonemes and visemes respectively, which are the basic units of speech sounds and mouth movements.
    To address audiovisual synchronization issues, we introduce the Phoneme-Viseme Alignment Mechanism \textbf{(PV-Align)}, which establishes robust cross-modal correspondences between phonemes and visemes.
    In addition, we build a Multilingual Talking Face Dataset \textbf{(MTFD)} comprising 12 diverse languages with 95.04 hours of high-quality videos for training and evaluating multilingual TFS performance.
    Extensive experiments demonstrate that MuEx achieves superior performance across all languages in MTFD and exhibits effective zero-shot generalization to unseen languages without additional training.
\end{abstract}



\keywords{Talking Face Synthesis, Phoneme-Viseme Alignment, MoE}

\maketitle

\section{Introduction}
Speech-driven TFS \cite{fan2026unisyncgeneralizablehighfidelitylip, tan2026flowportraitreinforcementlearningaudiodriven, li2026unitalkingunifiedaudiovideoframework, zhang2026exomnienabling3dfacial, lyu2026auheadrealisticemotionaltalking} has made remarkable progress in creating realistic facial animations with precise lip synchronization \cite{prajwal2020lip, zhang2023sadtalker, cui2025hallo, wang2025fantasytalking}. 
These advances enable applications that range from virtual assistants to content creation. However, current methods face severe limitations when processing non-English speech.

\begin{figure}[t]
\centering
\includegraphics[width=0.45\textwidth]{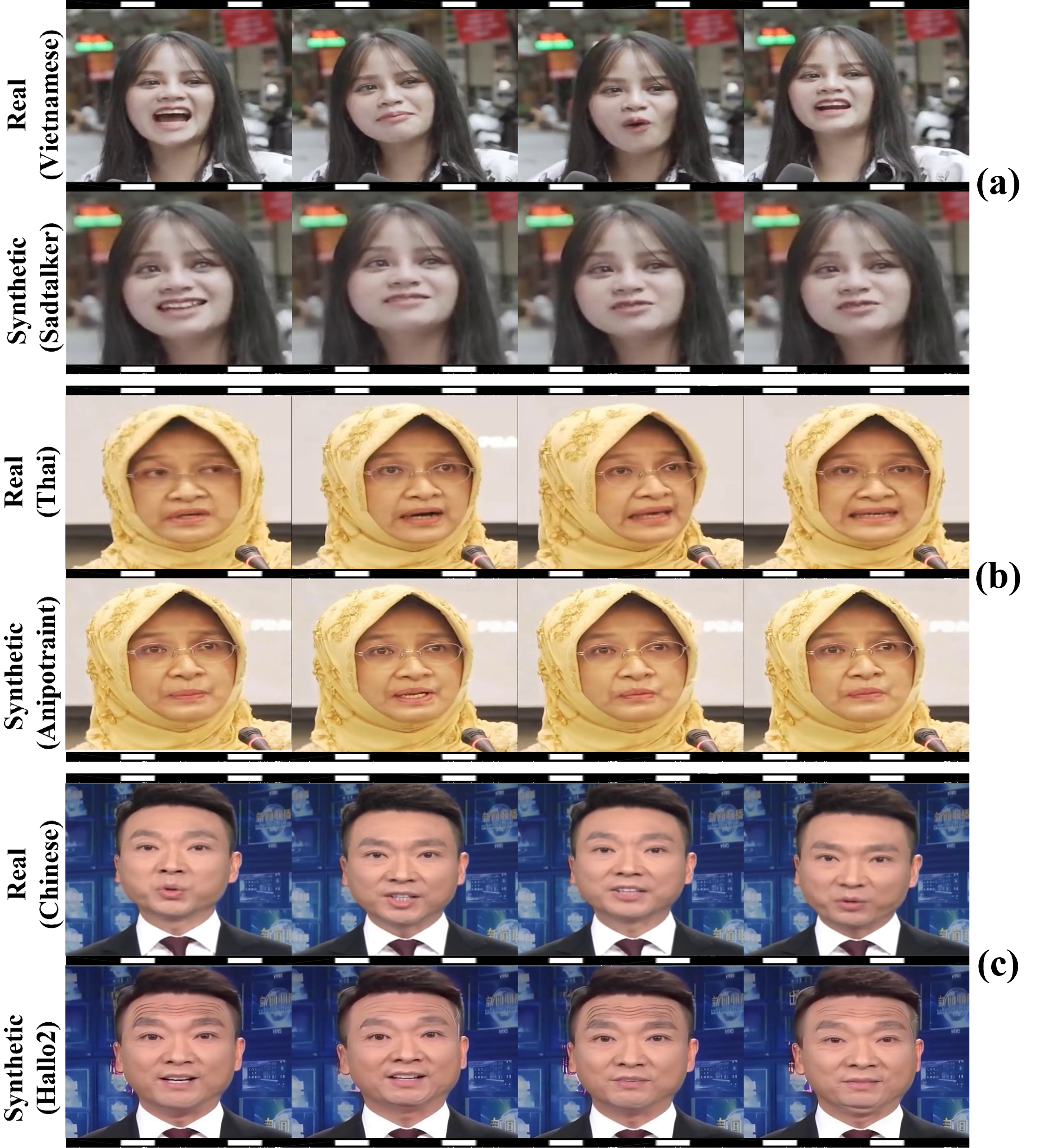}
\caption{Comparison of real and TFS videos on non-English languages reveals phoneme-viseme mismatches, rigid facial expressions, and audiovisual decoupling.}
\label{fig:motivation}
\end{figure}

As shown in Figure~\ref{fig:motivation}, three critical issues emerge when existing TFS models \cite{zhang2023sadtalker, wei2024aniportrait, cui2025hallo} process non-English languages. 
First, there are mismatches between phonemes (basic units of speech sounds) and visemes (corresponding mouth shapes) when models process new audio patterns, as shown in Figure~\ref{fig:motivation} (a).
Second, audiovisual decoupling leads to a more severe issue: clear speech audio does not produce the corresponding mouth movements, as shown in Figure~\ref{fig:motivation} (b).
Third, facial expressions become rigid, especially in tonal languages (Chinese), as shown in Figure~\ref{fig:motivation} (c).

The root cause of the above audiovisual mismatches is the result of two critical flaws in existing methods.
First, existing models are trained mainly on English datasets such as VoxCeleb \cite{nagrani2017voxceleb}, CelebV-HQ \cite{zhu2022celebv}, and HDTF \cite{zhang2021flow}, resulting in generation patterns optimized specifically for English audiovisual alignment.
When processing non-English speech with different phonetic structures, the pre-trained audiovisual alignment fails to generate accurate mouth movements from multilingual speech.
Second, existing approaches construct audiovisual alignment through end-to-end learning without modeling the underlying phonological principles.
Thus, these methods lack the ability to generalize to multilingual speech, resulting in incorrect mouth movements and poor audiovisual synchronization.

Therefore, we propose \textbf{MuEx}, a novel framework featuring a phoneme-guided mixture-of-experts \textbf{(PG-MoE)} architecture that employs phonemes and visemes as universal intermediaries to bridge the gap between audio and video modalities, achieving lifelike multilingual TFS.
Our approach is centered on two core innovations. 
First, we introduce the \textbf{PV-Align} mechanism, which constructs language-agnostic articulatory representations by clustering speech and visual features into phoneme and viseme prototypes, and then establishes robust cross-modal correspondences through prototype-level alignment.
This resolves phoneme-viseme mismatches by learning universal audiovisual correspondences that transcend linguistic boundaries.
Second, we design a \textbf{PG-MoE} that leverages these universal phoneme-viseme alignments for MoE routing. 
The system selects experts based on audiovisual similarities rather than language labels, enabling the model to choose appropriate processing pathways for multiple languages without explicit supervision.

Our framework supports speech-driven talking face generation across diverse languages without requiring explicit language labels, and demonstrates effective transfer to unseen languages in our experiments.

\vspace{0.5em}
\noindent\textbf{Our contributions include:}
\begin{itemize}
    \item We design a novel speech-driven face synthesis paradigm \textbf{(MuEx)} utilizing phonemes and visemes as intermediaries to bridge the gap between audio and video modalities, overcoming multilingual TFS generalization barriers and cross-modal synchronization challenges.
    \item We propose the Phoneme-Viseme Alignment \textbf{(PV-Align)} mechanism to establish robust cross-modal correspondences between phonemes and visemes, thereby enabling effective zero-shot generalization to unseen languages.
    \item We introduce a comprehensive Multilingual Talking Face Dataset \textbf{(MTFD)} comprising 12 languages with 95.04 hours of high-quality audiovisual content to train and evaluate multilingual TFS performance.
\end{itemize}

\section{Related Work}

\subsection{Audio-Driven Talking Head Generation}

Recent advances in speech-driven talking head generation \cite{wu2026asymmetrichierarchicalanchoringaudiovisual, wang2026joyavatarunlockinghighlyexpressive, chen2026justdubitvideodubbingjoint, flynn2026edityourselfaudiodrivengenerationmanipulation, zhang2026audiodriventalkingfacegeneration} have leveraged deep learning for improved realism. GAN-based methods such as Wav2Lip \cite{prajwal2020lip} and SadTalker \cite{zhang2023sadtalker} achieved strong lip-sync precision and control of facial expression. Diffusion-based approaches such as DiffTalk \cite{shen2023difftalk} and Hallo2 \cite{cui2025hallo} demonstrated the high-quality facial animation synthesis. Transformer architectures, including AniTalker \cite{peng2024anitalker} and EchoMimic \cite{chen2024echomimic} focused on temporal consistency and identity preservation. However, these methods primarily target single-language scenarios and struggle with multilingual generalization.

\subsection{Multilingual Speech and Learning}

Multilingual speech processing has made progress with self-supervised methods such as Wav2Vec 2.0 \cite{baevski2020wav2vec} and XLSR \cite{conneau2020unsupervised}, demonstrating multilingual transfer capabilities across diverse languages. Understanding phoneme-to-viseme correspondences is crucial for lip-sync generation, with research showing that languages exhibit distinct articulatory patterns and viseme distributions \cite{massaro1999perceiving}. multilingual transfer learning techniques, including few-shot learning \cite{finn2017model} and zero-shot generalization \cite{radford2021learning}, have shown promise in handling low-resource scenarios, while continual learning methods address catastrophic forgetting \cite{kirkpatrick2017overcoming}.


\subsection{Mixture of Experts}

Mixture of Experts (MoE) architectures \cite{wu2024multihead}  have proven effective in handling diverse tasks through sparse expert routing, as demonstrated in Switch Transformer \cite{fedus2022switch} and Vision MoE \cite{riquelme2021scaling}. For facial animation, discrete representation learning through vector quantization has gained attention. VQTalker \cite{liu2025vqtalker} introduced facial motion quantization for talking head generation, while methods such as VQ-VAE \cite{van2017neural} established foundations for discrete generative modeling.

Despite advances in TFS \cite{egin2026medontunifiedframework, ling2026synclipmaecontrastivemaskedpretraining, cao2024joyvasaportraitanimalimage, li2026joyavatarflashrealtimeinfiniteaudiodriven, ling2026synclipmaecontrastivemaskedpretraining}, existing methods struggle with multilingual scenarios, exhibiting phoneme-viseme mismatches and audiovisual decoupling when processing non-English speech. These English-centric approaches lack cross-language generalization and require complete retraining for different languages. Our MuEx framework addresses these limitations through PG-MoE and PV-Align, enabling robust multilingual synthesis with zero-shot transfer capabilities.

\section{Method}

\begin{figure*}[t]
    \centering
    \includegraphics[width=1\linewidth]{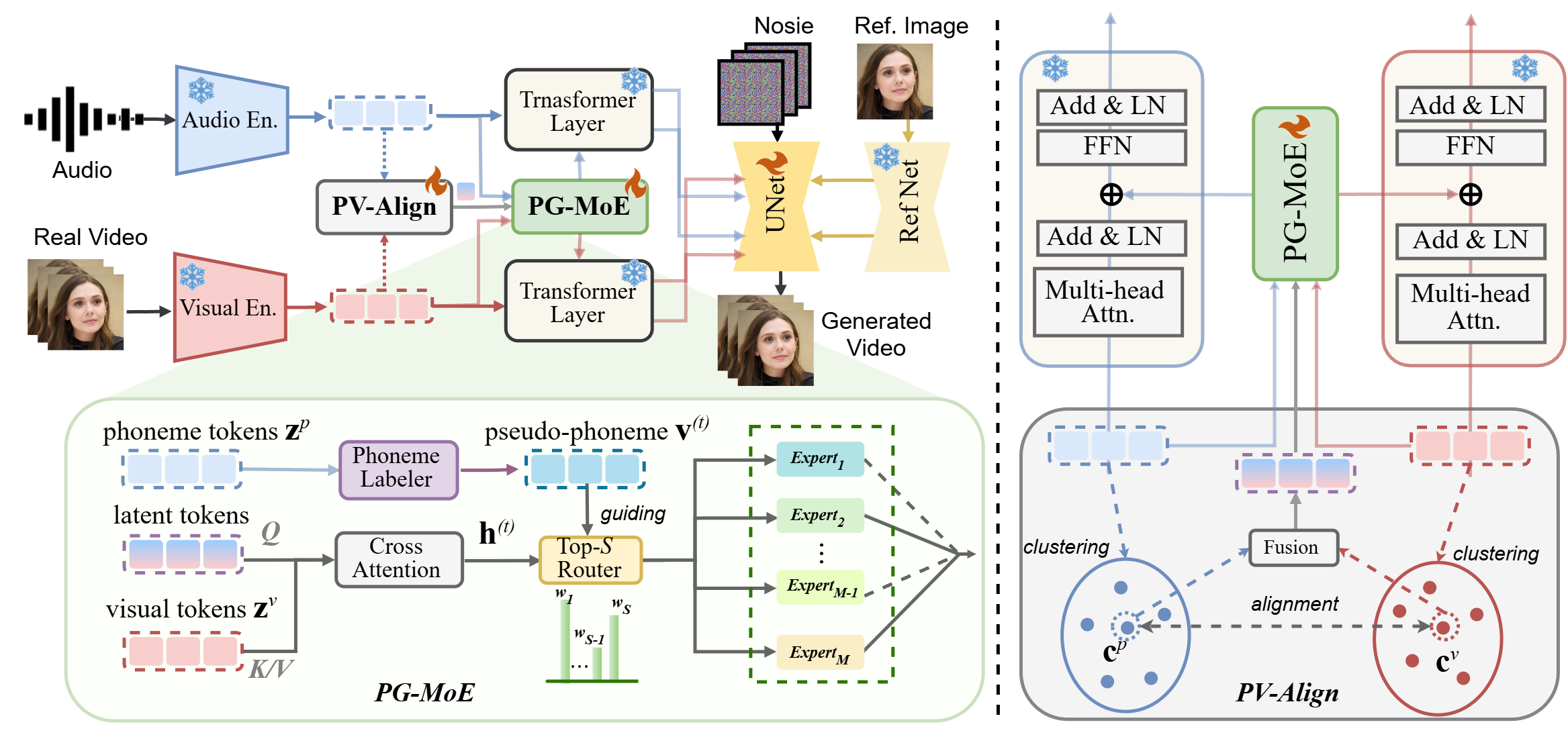}
    \caption{Framework. 
    The model learns universal viseme--phoneme prototypes and employs pseudo-phoneme guided expert routing 
    to enable cross-lingual speech-driven TFS.}
    \label{fig:framework}
\end{figure*}

\subsection{Overview}
Our goal is to improve cross-lingual generalization in speech-driven TFS. 
A key challenge lies in the variability of speech-to-lip correspondences across different languages, which often prevents models trained on one language from transferring effectively to others. 
To overcome this, we propose a framework that disentangles and aligns phoneme and viseme representations in a language-agnostic manner. 
As shown in Figure~\ref{fig:framework}, the pipeline consists of three major components. 
First, we cluster speech (phoneme-level) and video (viseme-level) features into prototypes, which act as universal anchors capturing articulatory units shared across languages. 
Second, we align phoneme and viseme prototypes by maximizing their mutual information while suppressing spurious correlations at the raw feature level, ensuring that the learned mapping reflects stable and transferable articulatory patterns. 
Third, we estimate pseudo-phonemes and use them to guide a sparse mixture-of-experts (MoE) router, which dynamically selects specialized expert modules to synthesize natural and synchronized lip movements in multiple languages. 
Together, these components form a unified architecture that builds a universal viseme--phoneme mapping and leverages it for cross-lingual speech-driven TFS.

\subsection{Phoneme--Viseme Alignment}

A central goal of our framework is to construct language-independent articulatory anchors that bridge the gap between audio and video modalities. 
To this end, we cluster speech features at the phoneme level and video features at the viseme level into $K$ prototypes, denoted as $\{\mathbf{c}^p_k\}_{k=1}^K$ and $\{\mathbf{c}^v_k\}_{k=1}^K$, respectively. 
We set $K$ close to the phoneme inventory size and initialize the prototypes with K-means++ for stable coverage of the feature space. 
To adapt to the evolving feature distribution during training, the prototypes are periodically updated every 10 epochs with modest computational overhead.

Given an input feature $\mathbf{z}^p_t$ or $\mathbf{z}^v_t$ at time step $t$, we assign it to the nearest prototype and obtain the corresponding discrete codes $\mathbf{q}^p_t$ and $\mathbf{q}^v_t$. 
Such prototype-based discretization provides interpretable articulatory units for both modalities. 
Although phoneme inventories differ across languages, many articulatory gestures, such as lip closure, mouth opening, and jaw movement, are shared across languages. 
Therefore, these prototypes serve as transferable anchors for cross-modal alignment.

\begin{figure}[htbp]
    \centering
    \includegraphics[width=0.8\linewidth]{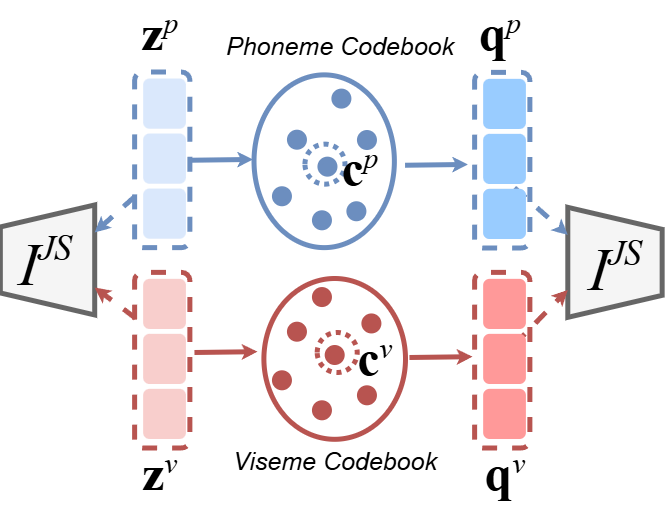}
    \caption{Illustration of PV-Align. Audio and video features are mapped into phoneme and viseme prototype codebooks, producing discrete assignments.}
    \label{fig:APVA}
\end{figure}

Figure~\ref{fig:APVA} illustrates the overall mechanism of PV-Align, where speech and visual features are first mapped to prototype codebooks and then aligned through the Jensen--Shannon MI objective.
Based on these discrete assignments, we define the phoneme--viseme alignment loss as
\begin{equation}
\mathcal{L}_{\mathrm{align}}
= -\, I^{\mathrm{JS}}\!\big(\mathbf{q}^{p}, \mathbf{q}^{v}\big)
\;+\; \lambda_{\mathrm{neg}}\, I^{\mathrm{JS}}\!\big(\mathbf{z}^{p}, \mathbf{z}^{v}\big),
\end{equation}
where $\mathbf{q}^{p}$ and $\mathbf{q}^{v}$ denote the phoneme and viseme prototype assignments, while $\mathbf{z}^{p}$ and $\mathbf{z}^{v}$ are the corresponding raw speech and visual features. 
As illustrated in Figure~\ref{fig:APVA}, this objective promotes meaningful cross-modal correspondences in the prototype space while suppressing noisy correlations in the raw feature space.

Specifically, the first term, $-I^{\mathrm{JS}}(\mathbf{q}^{p}, \mathbf{q}^{v})$, maximizes the Jensen--Shannon mutual information between matched phoneme and viseme assignments. 
This encourages speech and mouth-shape samples that co-occur in real data to be mapped to compatible prototype groups, making the learned clusters more consistent across modalities rather than being determined solely by within-modality similarity. 
The second term, $\lambda_{\mathrm{neg}} I^{\mathrm{JS}}(\mathbf{z}^{p}, \mathbf{z}^{v})$ serves as a \emph{weak regularizer} on the mutual information between raw speech and visual features, discouraging the model from over-relying on low-level cross-modal correlations. In particular, it aims to reduce \emph{nuisance correlations}, such as speaker-specific style, appearance-dependent patterns, and recording-condition artifacts, which are less transferable across languages.

Importantly, this regularization is not intended to remove content-relevant audio-visual consistency necessary for lip synchronization. Such consistency is preserved by the prototype-level alignment term together with the downstream generation objectives. In practice, $\lambda_{\mathrm{neg}}$ is set to a small value so that this term only provides a mild inductive bias rather than dominating the overall optimization.

To estimate mutual information in a unified manner, we adopt the Jensen--Shannon lower bound with a binary-classifier formulation. 
Let $\mathcal{P}$ denote the set of matched pairs and $\mathcal{N}$ the set of shuffled (mismatched) pairs. 
Using a lightweight discriminator $D_{\psi}$ that takes the concatenated feature pair $[\mathbf{x};\mathbf{y}]$ as input, the estimator is
\begin{equation}
\begin{aligned}
I^{\mathrm{JS}}(\mathbf{x}, \mathbf{y})
&= \mathbb{E}_{(\mathbf{x},\mathbf{y})\sim\mathcal{P}}
\!\left[\log \sigma\!\big(D_{\psi}([\mathbf{x};\mathbf{y}])\big)\right] \\
&\quad + \mathbb{E}_{(\mathbf{x},\mathbf{y})\sim\mathcal{N}}
\!\left[\log\!\big(1-\sigma\!\big(D_{\psi}([\mathbf{x};\mathbf{y}])\big)\big)\right],
\end{aligned}
\end{equation}
where $\sigma(\cdot)$ denotes the logistic sigmoid. 
This formulation encourages matched phoneme--viseme pairs to be distinguishable from mismatched ones, thereby improving the reliability of the learned cross-modal correspondence.


\subsection{Pseudo-Phoneme Guided Expert Routing}
Although prototype alignment provides universal anchors, the model still needs to adapt feature processing dynamically across languages.
To this end, we introduce a sparse mixture-of-experts (MoE) router guided by pseudo-phoneme distributions.
At each time step $t$, the router combines content features $\mathbf{h}^{(t)}$ with pseudo-phoneme features $\mathbf{v}^{(t)}$ to produce expert logits
\begin{equation}
a_i^{(t)}=\beta\,g_{\text{phon},i}(\mathbf{v}^{(t)})+(1-\beta)\,g_{\text{cont},i}(\mathbf{h}^{(t)}),
\quad i=1,\dots,M,
\end{equation}
where $\beta\in[0,1]$ balances phoneme-aware and content-aware routing.
The routing distribution is computed by
\begin{equation}
r_i^{(t)}=\frac{\exp(a_i^{(t)})}{\sum_{j=1}^{M}\exp(a_j^{(t)})}.
\end{equation}
The top-$S$ experts are selected for sparse computation, and their outputs are aggregated with normalized routing weights.

\begin{figure}[H]
    \centering
    \includegraphics[width=0.95\linewidth]{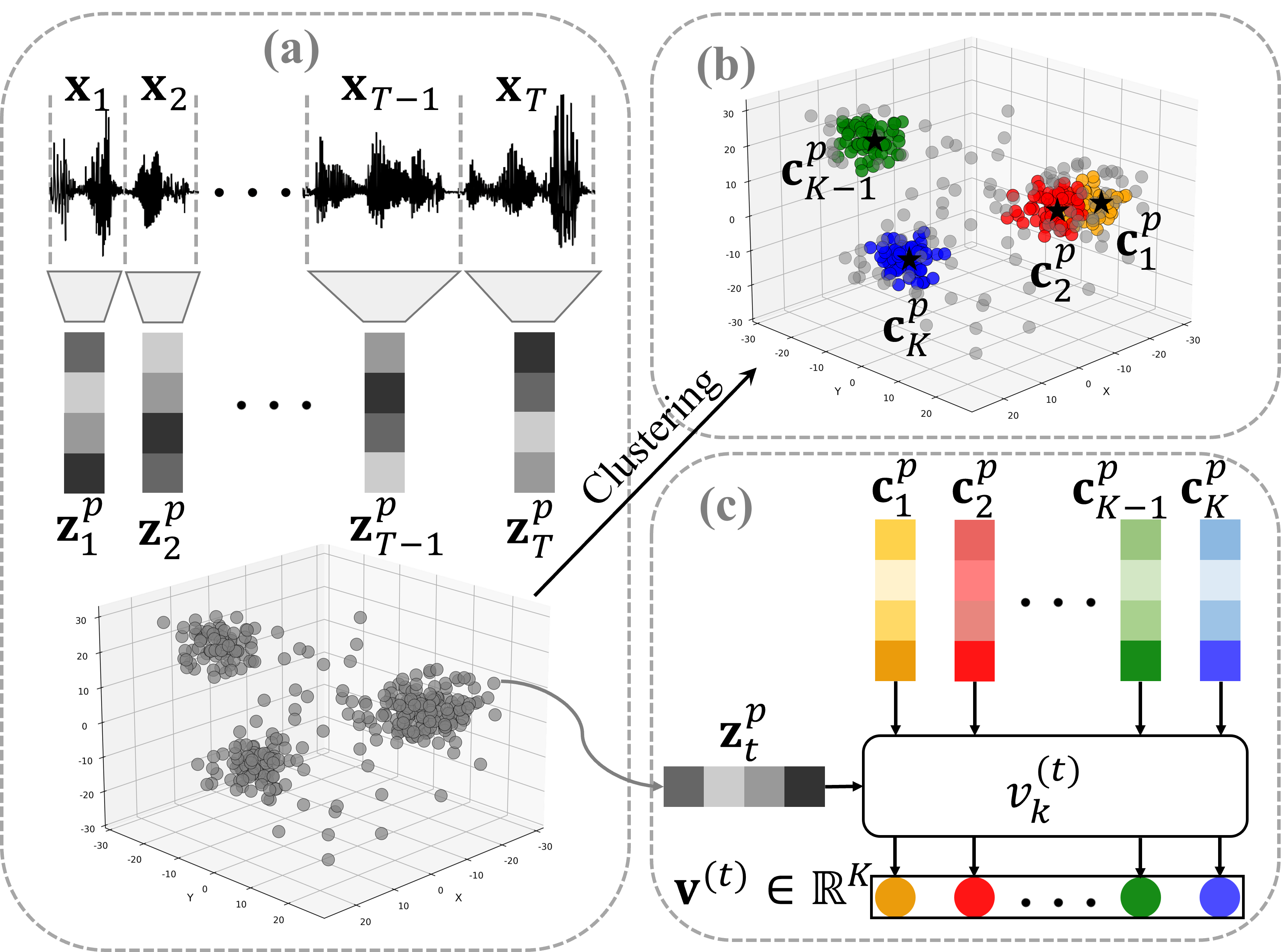}
    \caption{Mechanism of the Phoneme Labeler.}
    \label{fig:ClusterSkipgram}
\end{figure}

As illustrated in Figure~\ref{fig:ClusterSkipgram}, we compute a soft assignment over phoneme prototypes based on the similarity between the speech embedding $\mathbf{z}_t^p$ and the phoneme centroids $\{\mathbf{c}_k^p\}_{k=1}^K$:
\begin{equation}
v_k^{(t)}=
\frac{\exp\!\left(-\|\mathbf{z}_t^p-\mathbf{c}_k^p\|_2^2/\tau^2\right)}
{\sum_{j=1}^{K}\exp\!\left(-\|\mathbf{z}_t^p-\mathbf{c}_j^p\|_2^2/\tau^2\right)},
\quad k=1,\dots,K.
\end{equation}
The resulting pseudo-phoneme vector $\mathbf{v}^{(t)} = [v_1^{(t)}, v_2^{(t)}, \ldots, v_K^{(t)}]$ can be regarded as a phoneme-like latent representation that approximates the underlying acoustic-articulatory structure of true phonemes in a language-agnostic feature space. As a $K$-dimensional probability distribution, it measures the affinity between the input speech representation and each discovered phonetic prototype, allowing the model to capture shared cross-lingual phonetic structure without explicit language supervision. 

To make pseudo-phoneme supervision compatible with expert routing, we introduce a learnable projection matrix $\mathbf{A}\in\mathbb{R}^{K\times M}$ and define the pseudo-phoneme-guided expert target as
\begin{equation}
\tilde{\mathbf{r}}^{(t)}=\mathrm{softmax}(\mathbf{A}^{\top}\mathbf{v}^{(t)}).
\end{equation}

The routing loss is defined as
\begin{equation}
\mathcal{L}_{\text{router}}
=
-\frac{1}{T}\sum_{t=1}^{T}\sum_{i=1}^{M}\tilde{r}_i^{(t)}\log r_i^{(t)}
+\lambda_{\text{bal}}\sum_{i=1}^{M}\left(\bar{r}_i-\frac{1}{M}\right)^2,
\end{equation}
where
\begin{equation}
\bar{r}_i=\frac{1}{T}\sum_{t=1}^{T}r_i^{(t)}.
\end{equation}

The first term aligns the router output with the pseudo-phoneme-guided target distribution, while the second term encourages balanced expert utilization.
Using this simple routing objective, the router learns to activate experts associated with transferable cross-lingual articulatory patterns.

\subsection{Training Objective and Workflow}
Our model jointly optimizes prototype alignment, expert routing, and the TFS objective. 
The overall loss is defined as
\begin{equation}
\mathcal{L}
=
\lambda_{\mathrm{align}}\mathcal{L}_{\mathrm{align}}
+
\lambda_{\mathrm{router}}\mathcal{L}_{\mathrm{router}}
+
\lambda_{\mathrm{gen}}\mathcal{L}_{\mathrm{gen}},
\end{equation}
where $\mathcal{L}_{\text{align}}$ enforces universal phoneme--viseme mapping, 
$\mathcal{L}_{\text{router}}$ guides expert activation with pseudo-phoneme supervision, 
and $\mathcal{L}_{\text{gen}}$ drives the talking face generator. 

\textbf{Generation loss.}
To ensure both pixel fidelity and perceptual realism, we define
\begin{equation}
\mathcal{L}_{\text{gen}} = 
\lambda_{1}\|\hat{\mathbf{I}} - \mathbf{I}\|_{1} 
+ \lambda_{p}\|\phi(\hat{\mathbf{I}}) - \phi(\mathbf{I})\|_{2}^{2} 
+ \lambda_{t}\|\nabla_t \hat{\mathbf{I}} - \nabla_t \mathbf{I}\|_{1},
\end{equation}
where $\hat{\mathbf{I}}$ and $\mathbf{I}$ denote the generated and ground-truth video frames, 
$\phi(\cdot)$ extracts deep features from a pretrained perceptual network (VGG), 
and $\nabla_t$ denotes temporal frame differences. 
The first term enforces pixel-level accuracy, 
the second term preserves semantic and structural realism, 
and the third term improves temporal smoothness of lip movements. 
The weights $\lambda_{1}, \lambda_{p}, \lambda_{t}$ control the relative contributions of these objectives.

\textbf{Training workflow.}
During each training step, we (1) extract phoneme and viseme features from the input  and video, 
(2) assign features to prototypes and compute the alignment loss, 
(3) estimate pseudo-phonemes and apply MoE routing, 
(4) synthesize video frames and compute $\mathcal{L}_{\text{gen}}$, and 
(5) jointly update all modules by minimizing the total loss. 
This unified optimization enables the model to capture language-agnostic articulatory correspondences while generating realistic cross-lingual talking face videos.

\section{Experiments}
\subsection{Dataset Details}
\begin{figure}[H]
\centering
\includegraphics[width=0.80\linewidth]{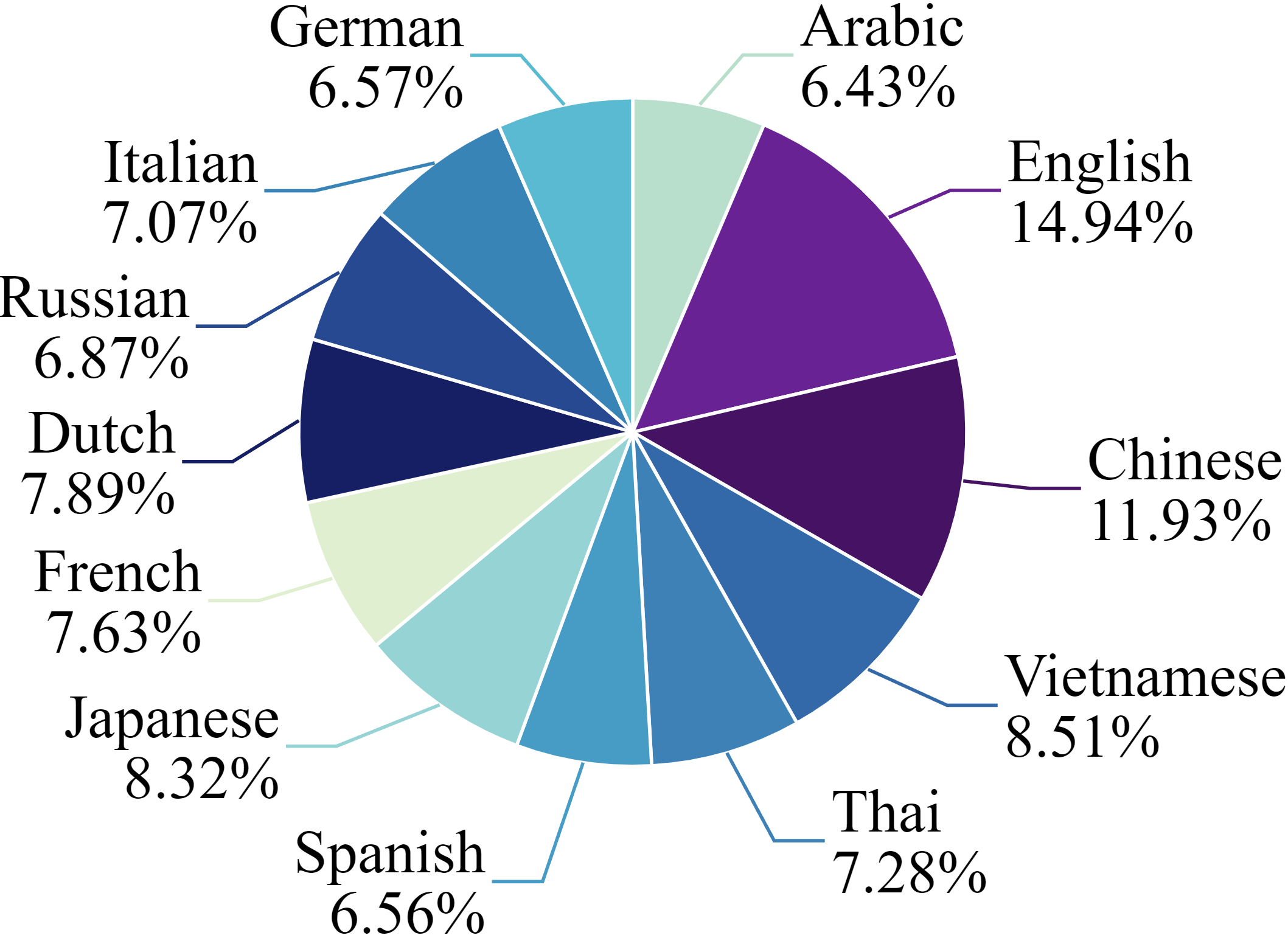}
\caption{Language distribution in MTFD.}
\label{fig:dataset-pie}
\end{figure}

We employ a two-stage training approach using complementary datasets. Initially, we trained on the HDTF \cite{zhang2021flow} dataset, which provides more than 15 hours of high-quality English talking face videos with 512×512 resolution with clear audiovisual synchronization.
For multilingual generalization, we constructed a multilingual talking face dataset (MTFD) by collecting news broadcasts and interview videos from 12 countries. As shown in figure~\ref{fig:dataset-pie}, the dataset contains 95.04 hours of high-quality content spanning diverse language families. Each video segment undergoes careful preprocessing to ensure a single-person 
appearance, frontal face orientation, and clean speech without background music or noise.
All videos are standardized to 512×512 resolution to maintain consistency with the HDTF baseline. The collection includes both tonal languages (Chinese, Thai, Vietnamese) and non-tonal languages, providing comprehensive coverage for evaluating multilingual performance. This data set addresses the lack of diverse multilingual datasets in TFS research, particularly for underrepresented languages like Arabic and Thai.

\subsection{Evaluation Metrics}

We used the Fréchet Video Distance (FVD)~\cite{unterthiner2019fvd} to quantify the distributional differences between the generated and real videos. Additionally, we adopt SyncNet~\cite{prajwal2020lip} and use the SyncNet confidence score (Sync-C) as a metric to assess speech-lip synchronization quality.
To better evaluate multilingual scenarios, we propose two novel metrics.

\subsubsection{Lip-Sync Error Distance (LSE-D)}

We propose LSE-D to measure geometric consistency between mouth movements in generated and ground truth videos. We extract 26 key lip landmarks using MediaPipe Face Mesh~\cite{kartynnik2019real} with standardized preprocessing for scale invariance. The frame-wise lip-sync error is:
\begin{equation}
\vspace{-5pt}
\text{LSE}_t = \frac{1}{26} \sum_{i=1}^{26} \| l^{real}_{t,i} - l^{gen}_{t,i} \|_2.
\vspace{-5pt}
\end{equation}
The overall LSE-D score is the temporal average: $\text{LSE-D} = \frac{1}{T} \sum_{t=1}^{T} \text{LSE}_t$, where lower values indicate better synchronization.

\subsubsection{Temporal Mouth Dynamics Correlation (TMDC)}

To capture fine-grained temporal consistency across linguistic patterns, we propose TMDC. For each frame $t$, we extract five scale-invariant mouth features from lip landmarks $\mathbf{L}_t$:
\begin{equation}
\mathbf{F}_t = \begin{bmatrix}
\|\mathbf{L}_{left} - \mathbf{L}_{right}\|_2 \\
\|\mathbf{L}_{top} - \mathbf{L}_{bottom}\|_2 \\
\text{ConvexHull}(\mathbf{L}_t) \\
\frac{\|\mathbf{L}_{left} - \mathbf{L}_{right}\|_2}{\|\mathbf{L}_{top} - \mathbf{L}_{bottom}\|_2 + \epsilon} \\
\frac{\|\mathbf{L}_{top} - \mathbf{L}_{bottom}\|_2}{\|\mathbf{L}_{left} - \mathbf{L}_{right}\|_2 + \epsilon}
\end{bmatrix} = \begin{bmatrix}
\text{Width}_t \\
\text{Height}_t \\
\text{Area}_t \\
\text{AspectRatio}_t \\
\text{Openness}_t
\end{bmatrix},
\end{equation}
where $\epsilon = 10^{-8}$ prevents the division from being zero.

For each feature $k$, we compute the Pearson correlation between real and generated sequences:
\begin{equation}
r_k = \frac{\sum_{t=1}^{T}(f^{real}_{k,t} - \bar{f}^{real}_k)(f^{gen}_{k,t} - \bar{f}^{gen}_k)}{\sqrt{\sum_{t=1}^{T}(f^{real}_{k,t} - \bar{f}^{real}_k)^2}\sqrt{\sum_{t=1}^{T}(f^{gen}_{k,t} - \bar{f}^{gen}_k)^2}}.
\end{equation}
The overall TMDC score is $\text{TMDC} = \frac{1}{5}\sum_{k=1}^{5} r_k$, where higher values indicate better temporal consistency.

\subsection{Experimental Results}

\begin{table}[htbp]
\centering
\renewcommand{\arraystretch}{1.2}
\resizebox{0.95\columnwidth}{!}{%
\begin{tabular}{l|c|cccc}
\toprule
\textbf{Method} & \textbf{Setting} & \textbf{FVD$\downarrow$} & \textbf{Sync-C$\uparrow$} & \textbf{LSE-D$\downarrow$} & \textbf{TMDC$\uparrow$} \\
\midrule
AniTalker   & Pretrained & 193.762 & 5.594 & 0.0557 & 0.543 \\
SadTalker   & Pretrained & 185.382 & 6.312 & 0.0498 & 0.646 \\
EchoMimic   & Pretrained & 172.987 & 6.081 & 0.0451 & \textcolor{blue}{0.712} \\
AniPortrait & Pretrained & 248.913 & 4.064 & 0.0487 & 0.641 \\
AniPortrait & Trained on MTFD & 236.844 & 4.193 & 0.0492 & 0.654 \\
Hallo2      & Pretrained & 161.942 & 7.450 & 0.0480 & 0.693 \\
Hallo2      & Trained on MTFD & \textcolor{blue}{158.625} & 7.469 & \textcolor{blue}{0.0447} & 0.705 \\
MEMO        & Pretrained & 158.842 & 7.453 & 0.0456 & 0.677 \\
MEMO        & Trained on MTFD & \textcolor{red}{156.918} & \textcolor{blue}{7.471} & 0.0453 & 0.698 \\
DICE-Talker   & Pretrained & 216.830 & 6.780 & 0.0471 & 0.710 \\
\cmidrule(lr){1-6}
\rowcolor{gray!30}
Ours & Trained on MTFD & 171.284 & \textcolor{red}{7.536} & \textcolor{red}{0.0437} & \textcolor{red}{0.756} \\
\midrule
Real video  & - & - & 8.914 & - & - \\
\bottomrule
\end{tabular}%
}
\caption{Comparison with other methods. ``Pretrained'' denotes direct inference using official released checkpoints on the MTFD test set, while ``Trained on MTFD'' denotes models adapted on the MTFD training split before evaluation.}
\label{tab:performance}
\end{table}

\begin{table*}[ht]
\centering
\resizebox{0.92\textwidth}{!}{
\begin{tabular}{@{}lcccccccc@{}}
\toprule
& \multicolumn{8}{c}{\textbf{Evaluation Metrics (Lip-Speech Synchronization / Teeth Visibility and Naturalness)}} \\
\cmidrule(lr){2-9}
\multirow{2}{*}{\textbf{Language}}
& \multicolumn{4}{c}{\textbf{MTFD-adapted models}}
& \multicolumn{4}{c}{\textbf{Off-the-shelf pretrained models}} \\
\cmidrule(lr){2-5} \cmidrule(lr){6-9}
& \textbf{MuEx}
& \textbf{MEMO}
& \textbf{Hallo2}
& \textbf{AniPortrait}
& \textbf{DICE-Talker}
& \textbf{EchoMimic}
& \textbf{SadTalker}
& \textbf{AniTalker} \\
\midrule

\multicolumn{9}{@{}l}{\textbf{In-MTFD}} \\
English    & \textcolor{red}{5.70/5.60} & \textcolor{blue}{5.45}/2.95 & 5.30/1.00 & 2.10/2.20 & 5.10/4.65 & 3.15/\textcolor{blue}{5.40} & 3.75/3.65 & 1.00/3.15 \\
Chinese    & \textcolor{red}{5.70/5.15} & 4.50/3.70 & 4.55/\textcolor{blue}{4.95} & 2.85/2.30 & 3.20/3.45 & \textcolor{blue}{4.65}/4.75 & 1.55/1.60 & 1.70/2.25 \\
Spanish    & \textcolor{red}{5.55/5.55} & \textcolor{blue}{5.15}/5.05 & 4.85/\textcolor{blue}{5.20} & 1.75/1.95 & 4.65/4.30 & 4.60/4.20 & 3.00/2.75 & 1.25/1.35 \\
French     & \textcolor{red}{5.60}/4.20 & \textcolor{blue}{5.20/4.55} & 4.20/3.85 & 2.85/\textcolor{red}{5.50} & 4.35/4.25 & 5.15/4.45 & 1.00/1.05 & 2.20/1.95 \\
German     & \textcolor{red}{5.65/5.35} & 4.35/4.55 & \textcolor{blue}{4.50/4.80} & 1.35/1.40 & 4.20/3.85 & 2.70/2.45 & 4.45/4.40 & 2.35/2.60 \\
Japanese   & \textcolor{red}{5.70/5.70} & 3.15/4.35 & 3.60/1.50 & 1.90/3.55 & 4.05/3.90 & \textcolor{blue}{4.80}/\textcolor{blue}{4.65} & 3.55/3.70 & 1.45/1.90 \\
Arabic     & \textcolor{red}{5.70/5.65} & \textcolor{blue}{4.50}/3.55 & 3.65/3.15 & 3.20/3.75 & 3.85/3.40 & 3.75/\textcolor{blue}{3.80} & 2.60/2.40 & 2.10/2.25 \\
Dutch      & \textcolor{red}{5.65}/\textcolor{blue}{5.00} & 4.85/4.25 & 4.25/3.20 & 2.70/\textcolor{red}{5.10} & 4.50/3.25 & \textcolor{blue}{4.95}/4.65 & 1.05/1.15 & 2.40/1.90 \\
Russian    & \textcolor{red}{5.80/5.60} & \textcolor{blue}{5.25}/4.00 & 4.55/\textcolor{blue}{4.55} & 1.90/3.05 & 3.45/3.30 & 4.55/3.95 & 2.45/2.05 & 1.75/1.80 \\
Italian    & \textcolor{red}{5.65/5.45} & 5.00/4.60 & \textcolor{blue}{5.15/4.80} & 3.75/3.20 & 5.10/3.95 & 3.45/4.55 & 1.15/1.70 & 1.85/1.30 \\
Thai       & \textcolor{red}{5.75/5.85} & 4.65/\textcolor{blue}{4.85} & 3.80/3.05 & \textcolor{blue}{4.75}/4.60 & 3.45/3.30 & 3.50/4.30 & 1.10/1.00 & 2.10/2.20 \\
Vietnamese & \textcolor{red}{5.65/5.30} & 3.45/3.60 & \textcolor{blue}{5.35}/4.55 & 3.55/3.00 & 4.30/3.55 & 3.45/\textcolor{blue}{5.15} & 1.55/1.30 & 1.45/1.70 \\

\midrule
\multicolumn{9}{@{}l}{\textbf{Out-of-MTFD}} \\
Korean     & \textcolor{red}{5.65/5.70} & 4.60/3.85 & 2.30/2.10 & \textcolor{blue}{4.80}/4.10 & 4.35/4.00 & 4.50/\textcolor{blue}{5.00} & 1.00/1.15 & 2.75/2.95 \\
Burmese    & \textcolor{red}{5.50/5.85} & 4.15/3.55 & \textcolor{blue}{5.05}/3.70 & 1.80/1.55 & 4.25/3.95 & 4.45/\textcolor{blue}{5.10} & 3.00/3.35 & 1.20/1.45 \\
Hindi      & \textcolor{red}{5.55/5.00} & 5.10/\textcolor{blue}{4.45} & 3.25/3.70 & 2.65/4.00 & 5.15/3.80 & \textcolor{blue}{5.30}/\textcolor{blue}{4.45} & 3.25/2.65 & 1.00/1.20 \\

\midrule[\heavyrulewidth]
\textbf{In-MTFD Avg}
& \textcolor{red}{5.68/5.37}
& \textcolor{blue}{4.63}/4.17
& 4.48/3.72
& 2.72/3.30
& 4.27/3.80
& 4.06/\textcolor{blue}{4.36}
& 2.27/2.23
& 1.80/2.03 \\
\textbf{Out-of-MTFD Avg}
& \textcolor{red}{5.57/5.52}
& 4.62/3.95
& 3.53/3.17
& 3.08/3.22
& 4.58/3.92
& \textcolor{blue}{4.75/4.85}
& 2.42/2.38
& 1.65/1.87 \\
\textbf{Overall Avg}
& \textcolor{red}{5.65/5.41}
& \textcolor{blue}{4.62}/4.12
& 4.25/3.56
& 2.82/3.28
& 4.33/3.82
& 4.25/\textcolor{blue}{4.49}
& 2.31/2.27
& 1.76/1.99 \\
\bottomrule
\end{tabular}
}
\caption{Human Evaluation Results: Comprehensive Weighted Ranking Scores. Higher is better. Red indicates the best result and blue indicates the second-best result.}
\label{tab:human_eval_comprehensive}
\end{table*}

\subsubsection{Quantitative Analysis}

We compare MuEx with seven representative state-of-the-art TFS methods, including AniTalker~\cite{peng2024anitalker}, SadTalker~\cite{zhang2023sadtalker}, EchoMimic~\cite{chen2024echomimic}, AniPortrait~\cite{wei2024aniportrait}, MEMO~\cite{zheng2024memo}, Hallo2 \cite{cui2025hallo}, and DICE-Talker~\cite{tan2025dicetalk}. 
To ensure a transparent and as fair as possible comparison, we adopt two evaluation settings for methods with publicly available training code. 
Specifically, AniPortrait, Hallo2, and MEMO are evaluated under both: (1) an off-the-shelf setting using their officially released pretrained checkpoints, and (2) a multilingual adaptation setting, where each model is fine-tuned on the MTFD training split using the same data partition, face cropping pipeline, audio preprocessing strategy, and evaluation protocol as MuEx. 
For AniTalker, SadTalker, EchoMimic, and DICE-Talker, whose full training pipelines are not publicly available, we report results obtained directly from their official pretrained checkpoints on the MTFD test set.

Table~\ref{tab:performance} summarizes the quantitative results. 
MuEx achieves the strongest overall performance on synchronization-oriented metrics, obtaining the highest Sync-C score of 7.536, the lowest LSE-D of 0.0437, and the best TMDC of 0.756. 
Although its FVD of 171.284 is not the best among all methods, it remains competitive and indicates that MuEx preserves good visual quality while substantially improving audiovisual alignment. 
In contrast, MEMO fine-tuned on MTFD achieves the best FVD (156.918), but still falls behind MuEx on all synchronization-related metrics, suggesting that stronger global video realism does not necessarily translate into more accurate lip synchronization or better temporal mouth dynamics.

Among the adapted baselines, Hallo2 is the strongest competitor in terms of balanced performance. 
Compared with Hallo2 fine-tuned on MTFD, MuEx improves Sync-C from 7.469 to 7.536, reduces LSE-D from 0.0447 to 0.0437, and increases TMDC from 0.705 to 0.756. 
These gains correspond to relative improvements of 0.9\%, 2.2\%, and 7.2\%, respectively, demonstrating that MuEx produces more accurate and temporally coherent lip movements. 
Compared with MEMO fine-tuned on MTFD, MuEx further improves Sync-C by 0.87\%, reduces LSE-D by 3.5\%, and improves TMDC by 8.3\%, highlighting the advantage of our method in modeling cross-lingual phoneme-viseme correspondences even when the overall perceptual realism is not the absolute best in terms of FVD.

MuEx also consistently outperforms off-the-shelf pretrained baselines such as SadTalker, EchoMimic, and DICE-Talker on all synchronization-centered metrics. 
For example, compared with SadTalker, MuEx improves Sync-C from 6.312 to 7.536, reduces LSE-D from 0.0498 to 0.0437, and increases TMDC from 0.646 to 0.756. 
A similar trend can be observed against EchoMimic and DICE-Talker, indicating that existing English-centric pretrained models do not generalize sufficiently well to multilingual scenarios, whereas MuEx can better maintain speech-lip consistency across languages.

Overall, these results verify the effectiveness of the proposed PG-MoE and PV-Align modules. 
The clear gains on Sync-C, LSE-D, and TMDC demonstrate that MuEx is particularly effective at improving multilingual lip synchronization and temporal mouth motion consistency, which are the core challenges in cross-lingual talking face synthesis.

\begin{figure*}[ht]
\centering
\includegraphics[width=0.9\textwidth]{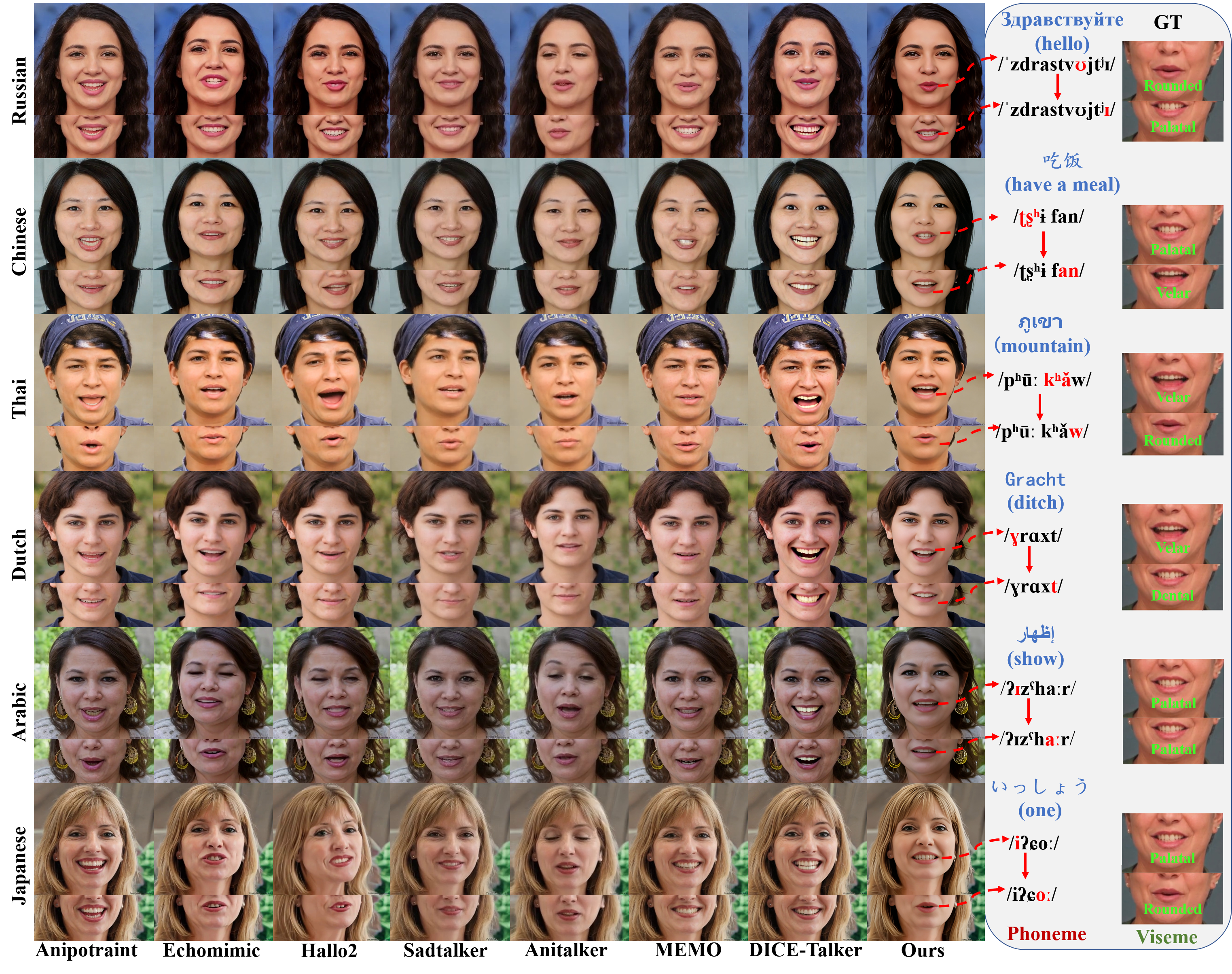}
\caption{Qualitative comparison across multiple languages. We transcribe test words into International Phonetic Alphabet (\textbf{IPA}) notation and extract representative video frames corresponding to specific phoneme pronunciations for comparison with the ground-truth mouth shapes (\textbf{GT}). Here, \textbf{phonemes} denote the basic units of speech sounds, while \textbf{visemes} denote their corresponding visual mouth configurations. MuEx produces more accurate viseme patterns and more natural articulatory transitions across languages. Hallo2, AniPortrait, and MEMO are shown using models fine-tuned on MTFD, while AniTalker, SadTalker, EchoMimic, and DICE-Talker are evaluated using their official pretrained checkpoints on the MTFD test set.}
\label{fig:visual-results}
\end{figure*}

\begin{table}[htbp]
\centering
\renewcommand{\arraystretch}{1.2}
\resizebox{0.95\columnwidth}{!}{%
\begin{tabular}{lcccc}
\toprule
\textbf{Configuration} & \textbf{FVD$\downarrow$} & \textbf{Sync-C$\uparrow$} & \textbf{LSE-D$\downarrow$} & \textbf{TMDC$\uparrow$} \\
\midrule
\makecell[l]{Baseline \\ (w/o PG-MoE)} & 171.547 & 6.921 & 0.0612 & 0.583 \\
\cmidrule(lr){1-5}
\makecell[l]{+ PG-MoE \\ (w/o PV-Align)} & 171.392 & 7.089 & 0.0551 & 0.629 \\
\cmidrule(lr){1-5}
\makecell[l]{+ PG-MoE \\ (w/o Phoneme Labeler)}   & 171.294 & 7.284 & 0.0481 & 0.686 \\
\cmidrule(lr){1-5}
\rowcolor{gray!30}
\textbf{MuEx (Full)} & \textbf{171.284} & \textbf{7.536} & \textbf{0.0437} & \textbf{0.756} \\
\bottomrule
\end{tabular}%
}
\caption{Ablation study on key components}
\label{tab:ablation}
\end{table}

\subsubsection{Qualitative Analysis}
We conducted a comprehensive human evaluation involving 300 participants across 15 language groups. Each group consisted of 20 speakers who evaluated 8 methods on two dimensions: (1) lip--speech synchronization consistency and (2) teeth visibility clarity and naturalness. Participants ranked the methods from 1st to 8th place, and weighted average scores were computed accordingly, with higher scores indicating better perceptual quality. The evaluation covered 12 in-MTFD languages and 3 unseen out-of-MTFD languages (Korean, Burmese, and Hindi), enabling assessment of both multilingual generation quality and zero-shot generalization.

\begin{figure*}[ht]
\centering
\includegraphics[width=0.9\textwidth]{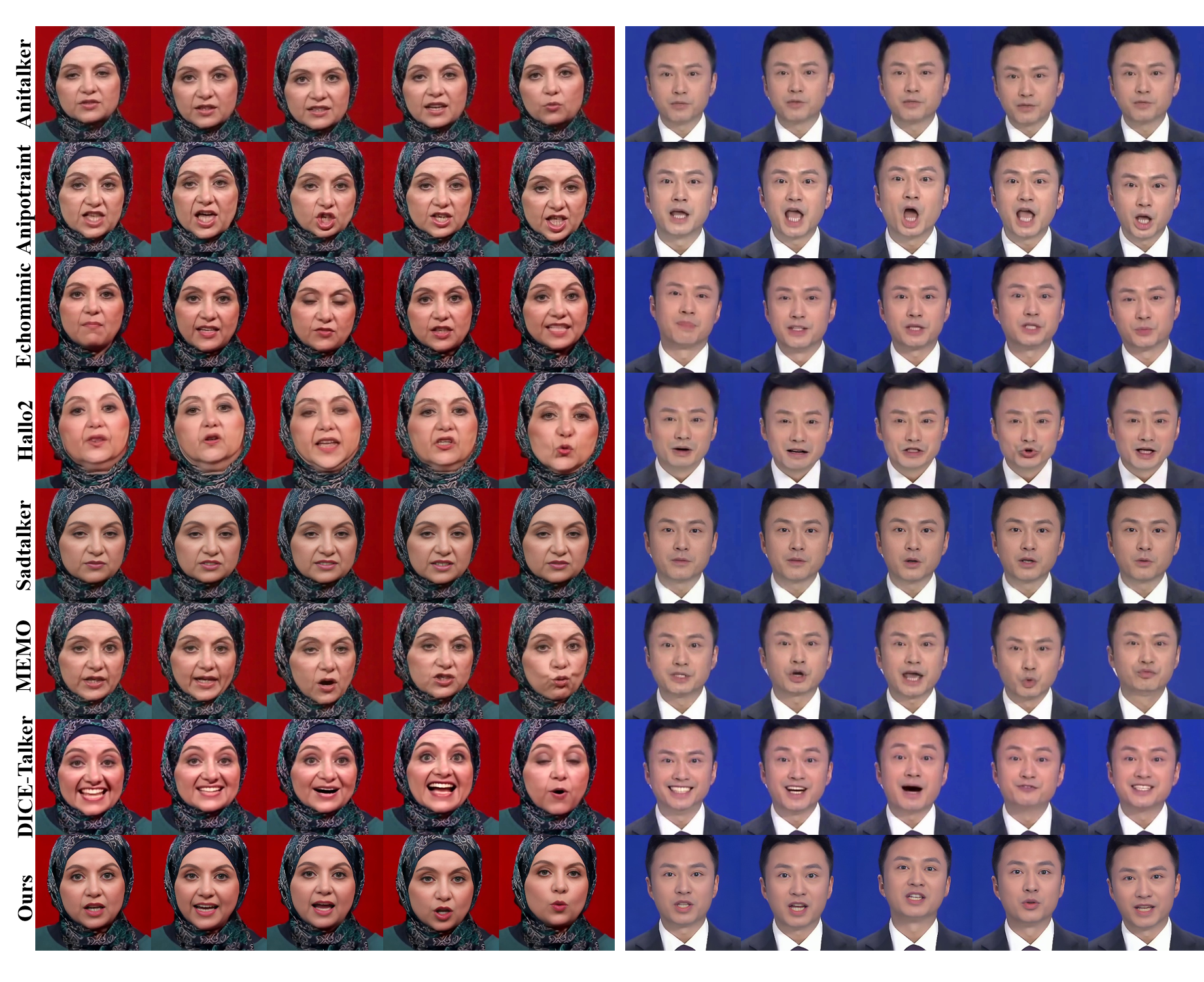}
\caption{Qualitative comparison on Arabic (left) and Chinese (right). MuEx generates more accurate mouth shapes and more natural facial expressions across distinct phonetic contexts, demonstrating stronger cross-lingual phoneme-viseme alignment.}
\label{fig:visual-results-2}
\end{figure*}

As illustrated in Figures~\ref{fig:visual-results} and \ref{fig:visual-results-2}, \textsc{MuEx} generates more accurate mouth shapes and smoother temporal dynamics across representative languages, showing clearer phoneme-to-viseme correspondence than the compared baselines. The human evaluation results in Table~\ref{tab:human_eval_comprehensive} are consistent with these visual observations. \textsc{MuEx} achieves the best overall average scores among all methods, reaching 5.65 in lip--speech synchronization and 5.41 in teeth naturalness, and remains the top-performing method on both the in-MTFD subset (5.68/5.37) and the out-of-MTFD subset (5.57/5.52).

The advantage of \textsc{MuEx} is particularly evident in tonal languages such as Chinese, Thai, and Vietnamese, where it obtains 5.70/5.15, 5.75/5.85, and 5.65/5.30, respectively. Moreover, its strong performance on Korean, Burmese, and Hindi indicates that the proposed phoneme-viseme alignment and expert routing strategy generalize effectively to unseen languages without additional training.

\subsection{Ablation Studies}

\subsubsection{Core Component Analysis}
Table~\ref{tab:ablation} presents a comprehensive ablation study evaluating each key component of our MuEx. The baseline model without PG-MoE shows substantial deficiencies in multilingual synchronization (LSE-D: 0.0612, TMDC: 0.583), confirming that specialized expert routing specifically targets multilingual synchronization challenges. Adding PG-MoE modules without PV-Align provides notable improvements (LSE-D: 0.0551, TMDC: 0.629), demonstrating MoE effectiveness for multilingual variations. The configuration with PG-MoE but without phoneme-guided routing achieves substantial improvements (LSE-D: 0.0481, TMDC: 0.686) through content-based expert selection, yet lacks the language-agnostic inductive bias for optimal cross-linguistic generalization. The complete MuEx delivers optimal performance across all metrics, eliminating explicit language supervision while enabling superior multilingual generalization. Progressive improvements demonstrate framework effectiveness: LSE-D improves by 28.6\% and TMDC by 29.7\% compared to baseline, validating the superiority of acoustic-articulatory principles for multilingual lip synchronization.

\subsubsection{Hyperparameter Selection}

{
\setlength{\textfloatsep}{4pt}
\setlength{\intextsep}{4pt}

\begin{table}[htbp]
\centering
\resizebox{1\columnwidth}{!}{%
\begin{tabular}{ccc|cccc|cc}
\toprule
\multicolumn{3}{c|}{\textbf{Hyperparameters}} & \multicolumn{4}{c|}{\textbf{Performance Metrics}} & \multicolumn{2}{c}{\textbf{Efficiency Metrics}} \\
\cmidrule(lr){1-3} \cmidrule(lr){4-7} \cmidrule(lr){8-9}
\textbf{S} & \textbf{M} & \textbf{K} & \textbf{FVD$\downarrow$} & \textbf{Sync-C$\uparrow$} & \textbf{LSE-D$\downarrow$} & \textbf{TMDC$\uparrow$} & \textbf{Memory(GB)$\downarrow$} & \textbf{FPS$\uparrow$} \\
\midrule
1 & 2 & 36 & 182.536 & 6.874 & 0.0534 & 0.641 & \textbf{1.7} & \textbf{31.2} \\
1 & 3 & 40 & 179.482 & 7.091 & 0.0501 & 0.679 & 2.0 & 28.7 \\
1 & 4 & 44 & 176.395 & 7.251 & 0.0476 & 0.703 & 2.5 & 25.8 \\
2 & 3 & 40 & 174.826 & 7.329 & 0.0463 & 0.721 & 2.2 & 26.4 \\
\rowcolor{gray!30} 
2 & 4 & 44 & \textbf{171.284} & \textbf{7.536} & \textbf{0.0437} & \textbf{0.756} & 2.8 & 22.1 \\
2 & 5 & 48 & 173.214 & 7.489 & 0.0452 & 0.734 & 3.3 & 19.5 \\
3 & 4 & 44 & 174.157 & 7.467 & 0.0458 & 0.742 & 3.0 & 20.8 \\
3 & 6 & 52 & 176.739 & 7.423 & 0.0471 & 0.718 & 4.1 & 16.3 \\
\bottomrule
\end{tabular}%
}
\caption{Trade-off Analysis of Hyperparameter Settings}
\label{tab:HP}
\end{table}
}

Table~\ref{tab:HP} demonstrates a clear performance efficiency trade-off across different hyperparameter configurations. The optimal configuration $(S=2, M=4, K=44)$ achieves the best overall performance (FVD: 171.284, Sync-C: 7.536, LSE-D: 0.0437, TMDC: 0.756) while maintaining reasonable computational efficiency (22.1 FPS, 2.8 GB memory). Lightweight configurations like $(S=1, M=2, K=36)$ maximize inference speed (31.2 FPS) and minimize memory usage (1.7 GB) at the cost of significant performance degradation. Notably, performance gains are not monotonic - the $(S=2, M=5, K=48)$ configuration underperforms despite higher complexity, while the heaviest configuration $(S=3, M=6, K=52)$ shows performance regression, indicating that excessive expert capacity can lead to training instability and suboptimal convergence, emphasizing the importance of balanced hyperparameter selection.

\section{Conclusion}

We propose MuEx, a novel framework that addresses the critical limitations of existing TFS methods in multilingual scenarios. Our approach introduces two key innovations: a PV-Align mechanism that creates language-agnostic cross-modal correspondences, and a PG-MoE that enables intelligent expert routing based on audiovisual similarities. Through comprehensive evaluation on our newly constructed MTFD with 12 languages, MuEx demonstrates superior performance in multilingual audiovisual synchronization while maintaining English capabilities and achieving effective zero-shot generalization to unseen languages. This work establishes a new paradigm for multilingual TFS that transcends language boundaries through universal phoneme-viseme principles.

\clearpage
\bibliographystyle{ACM-Reference-Format}
\bibliography{sample-base}

\appendix









\end{document}